%% file: eccv2022submission.tex
\documentclass[runningheads]{llncs}
\usepackage{graphicx}

\usepackage{tikz}
\usepackage{comment}
\usepackage{amsmath,amssymb} 
\usepackage{color}

\usepackage[accsupp]{axessibility}  

\usepackage{mmstyle}
\usepackage{xcolor}
\usepackage{graphicx}
\usepackage{float}
\usepackage{amsmath}
\usepackage{multirow}
\usepackage{booktabs}
\usepackage{colortbl}
\usepackage{pifont}
\usepackage[caption=false]{subfig}
\usepackage{makecell}
\usepackage{marvosym}
\usepackage[font=small]{caption}
\usepackage[pagebackref=true,breaklinks=true,colorlinks,bookmarks=false]{hyperref}
\usepackage{adjustbox}

\newcommand{\R}[1]{\textcolor{red}{#1}}
\newcommand{\B}[1]{\textcolor{blue}{#1}}
\newcommand{\bd}[1]{\textbf{#1}}

\newcommand\hfilll{\hspace{0pt plus 1filll}}

\newcommand{\cmark}{\ding{51}}
\newcommand{\xmark}{\ding{55}}

\begin{document}
\pagestyle{headings}
\mainmatter
\def\ECCVSubNumber{915}  

\title{MINI: Mining Implicit Novel Instances \\ for Few-Shot Object Detection} 

\titlerunning{MINI: Mining Implicit Novel Instances} 
\authorrunning{Y. Cao, J. Wang, Y. Lin, D. Lin} 
\author{
  Yuhang Cao$^{1}$ \quad Jiaqi Wang$^{1, 2}{\textsuperscript{\Letter}}$\thanks{\textsuperscript{\Letter}Corresponding author.} \quad Yiqi Lin$^3$ \quad Dahua Lin$^{1,2}$ \vspace{5pt}
}

\institute{
$^1$CUHK-SenseTime Joint Lab, The Chinese University of Hong Kong \\
$^2$Shanghai AI Laboratory \quad $^3$SenseTime Research \\
{\tt\small \{cy020,dhlin\}@ie.cuhk.edu.hk}\hspace{10pt} \\
{\tt\small wjqdev@gmail.com} \hspace{10pt}
{\tt\small linyiqi@sensetime.com} \hspace{10pt}
\\
}

\maketitle

\input{sections/abstract}
\input{sections/introduction}
\input{sections/related_work}
\input{sections/method}
\input{sections/experiments}
\input{sections/conclusion}
\input{sections/appendix.tex}

%

%
%
\bibliographystyle{splncs04}
\bibliography{egbib}
\end{document}

%% file: sections/abstract.tex
\begin{abstract}
    Learning from a few training samples is a desirable ability of an object detector, inspiring the explorations of Few-Shot Object Detection (FSOD).
    Most existing approaches employ a pretrain-transfer paradigm. The model is first pre-trained on base classes with abundant data and then transferred to novel classes with a few annotated samples. 
    Despite the substantial progress, the FSOD performance is still far behind satisfactory.
    During pre-training, due to the co-occurrence between base and novel classes, the model is learned to treat the co-occurred novel classes as backgrounds.
    During transferring, given scarce samples of novel classes, the model suffers from learning discriminative features to distinguish novel instances from backgrounds and base classes.
    To overcome the obstacles, we propose a novel framework, \emph{Mining Implicit Novel Instances (MINI)}, to mine the implicit novel instances as auxiliary training samples, which widely exist in abundant base data but are not annotated.
    MINI comprises an offline mining mechanism and an online mining mechanism.
    The offline mining mechanism leverages a self-supervised discriminative model to collaboratively mine implicit novel instances with a trained FSOD network.
    Taking the mined novel instances as auxiliary training samples, the online mining mechanism takes a teacher-student framework to simultaneously update the FSOD network and the mined implicit novel instances on the fly.
    Extensive experiments on PASCAL VOC and MS-COCO datasets show MINI achieves new state-of-the-art performance on any shot and split. The significant performance improvements demonstrate the superiority of our method.

    \keywords{Few-Shot Object Detection, Mining Implicit Novel Instances}
\end{abstract}

%% file: sections/introduction.tex
\section{Introduction}
\label{sec:intro}




\begin{figure}[t]
    \centering
    \includegraphics[width=1.0\linewidth]{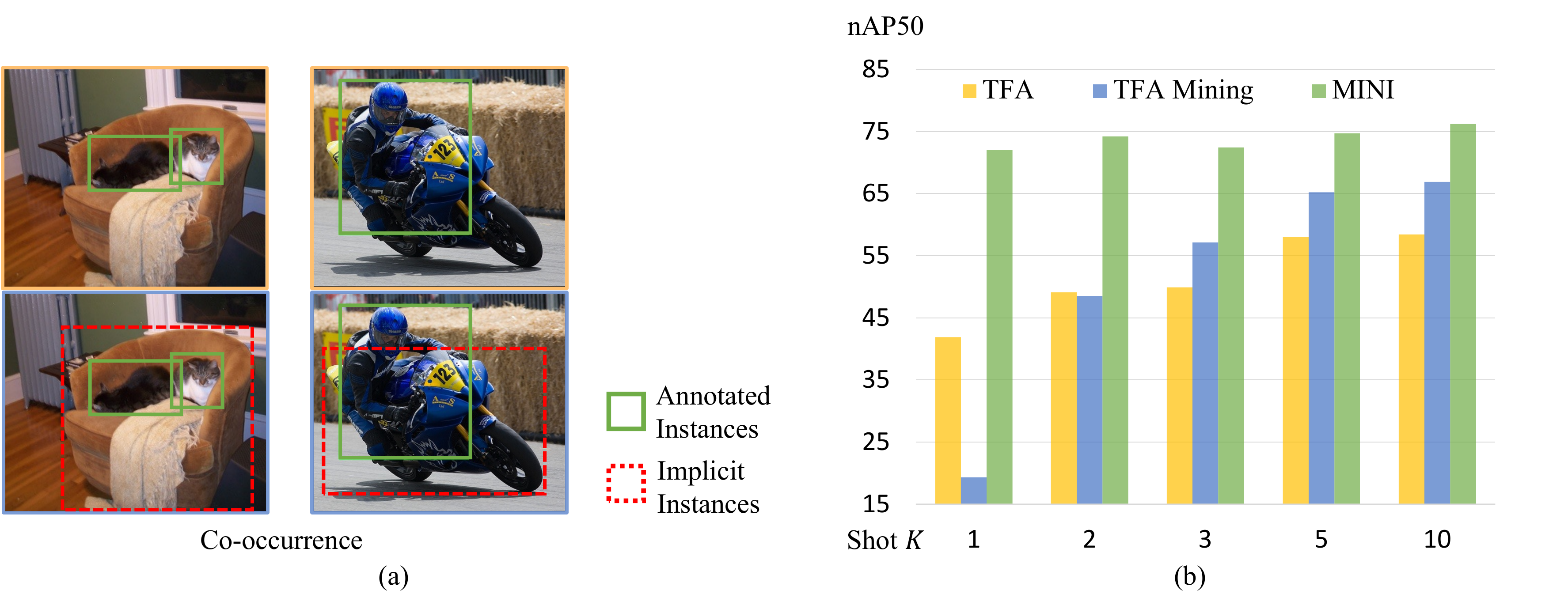}
    \vspace{-5mm}
    \caption{
        (a) Left figure demonstrates annotated instances of base classes (in green) and implicit instances of novel classes (in red) of FSOD datasets,
        where co-occurrence, \eg, a ``cat'' lies on a ``sofa'', a ``person'' rides a ``motor'', is widely existed.
        (b) Right figure compares the performance of different FSOD methods on PASCAL VOC dataset. The TFA~\cite{wang2020few} which is pre-trained on base classes is learned to treat implicit novel instances as backgrounds, resulting in unsatisfactory performance.
        Simply applying TFA to mine the implicit novel instances (TFA Mining) and re-train a detector with these instances is a straightforward solution. However, its performance is limited by the inaccurate initial detection results of TFA on low shots.
        The proposed \emph{MINI} can better mine these instances and significantly boost performance
    }
    \label{fig:teaser}
    \vspace{-6mm}
\end{figure}



Object detection aims to classify and localize objects, which receives remarkable progress in recent years~\cite{ren2015faster,he2017mask,cai18cascadercnn}.
However, the strong performance heavily relies on a large number of labeled training samples, which requires extensive labeling for each object and is expensive to acquire.
On the contrary, humans can recognize novel classes with the aid of only a few annotated samples,
which is a desirable ability object detectors should have. Thus great interests have been invoked to explore Few-Shot Object Detection (FSOD), which aims to train an object detector for novel classes with the help of abundant data on base classes and few shot samples on novel classes.

Current FSOD methods mostly follow a \emph{pretrain-transfer} paradigm. Specifically, it first pre-trains the object detector on the base classes with abundant data to attain general representation ability. And the pre-trained model is then transferred with few shot training samples to detect novel classes. Due to the limited number of novel samples, most parameters of the FSOD model are frozen when being transferred to preserve the pre-trained knowledge and prevent overfitting. 
Although various methods have been proposed following this paradigm, including meta-learning~\cite{yan2019meta,xiao2020few,fan2020few}, metric learning~\cite{karlinsky2019repmet,li2021beyond}, and fine-tuning~\cite{wang2020few,cao2021few,sun2021fsce}, their performance is still far behind satisfactory on benchmark datasets.

In this paper, we reveal that the performance of the current FSOD methods is heavily hindered by two aspects. 
First, the scarce novel samples fail to provide sufficient diversity of novel classes, making FSOD models tend to overfit to these few shot samples.
Second, due to the co-occurrence between base and novel classes on benchmark datasets, the object detector pre-trained on base classes is learned to treat the co-occurred novel instances as backgrounds. The classification bias is hard to be eliminated with frozen most parameters of the pre-trained model during transferring.


Motivated by the observations, this paper proposes to tackle this problem by mining the implicit novel instances, which widely exist in abundant data of base classes but are not annotated in FSOD datasets as in Fig.~\ref{fig:teaser}(a). By discovering these implicit novel instances and taking them as extra training samples, we can optimize all parameters of the FSOD model to solve the two mentioned obstacles at once. 
On the one hand, the enriched training sample of novel classes enhance the representation ability to discriminate novel classes from other classes. On the other hand, it effectively mitigates the classification confusion between backgrounds and novel instances. 

To achieve this goal, a straightforward solution is to directly adopt a FSOD model to discover these implicit novel instances. 
However, this simple design heavily relies on the initial performance of the FSOD model, leading to unsatisfactory performance, especially in low shots scenarios. 
Moreover, it lacks a mechanism to upgrade the discovered novel instances as the FSOD model improves, which hinders further performance improvement.

Towards the aforementioned drawbacks, this paper proposes a framework called \emph{Mining Implicit Novel Instances (MINI)} which mines the implicit novel instances with an offline mining mechanism and an online mining mechanism. 
Specifically, a FSOD model, \eg, TFA~\cite{wang2020few}, is firstly trained to discover initial implicit novel instances, as in Fig.~\ref{fig:teaser}(b).
The offline mining mechanism leverages a self-supervised discriminative model to calibrate classification confidences of these discovered novel instances.
During training, taking the offline mined implicit novel instances as auxiliary training samples, the online mining mechanism takes a teacher-student framework to simultaneously update the parameters of the FSOD network and the mined implicit novel instances on the fly. 


We conduct extensive experiments on Pascal VOC~\cite{Everingham10} and MS COCO~\cite{lin2014microsoft} benchmarks, and achieve new SOTA performance for all settings.
Concretely, we improve the current SOTA performance (novel AP50) by 18.4, 16.7, 10.9, 10.6, 12.8 and 19.3, 15.5, 15.3, 8.8, 13.5 and 16.6, 15.6, 11.7, 11.9, 10.8 for $K$=1, 2, 3, 5, 10 on novel split 1, 2 and 3, respectively.
Even on the challenging COCO split, we push the limit of the envelope performance (novel mAP) by 3.3 and 4.7 for $K=$ 10 and 30, respectively.
The significant performance gain demonstrates the effectiveness of the proposed \emph{MINI}.


%% file: sections/related_work.tex
\section{Related Work}
\label{sec:related}

\subsection{Few-Shot Object Detection}
Few-Shot Object Detection(FSOD) aims to detect novel concepts given abundant base data and limited novel data.
One main line of FSOD methods is meta-learning based approaches~\cite{karlinsky2019repmet,kang2019few,yan2019meta,xiao2020few,fan2020few,li2021beyond,zhang2021accurate,hu2021dense}.
FSRW~\cite{kang2019few} and Meta R-CNN~\cite{yan2019meta} introduce feature re-weighting to one-stage and two-stage detection methods, respectively. 
Meta-Det~\cite{wang2019meta} disentangles the learning of category-specific and category-agnostic components. 
FSIW~\cite{xiao2020few} improves FSRW~\cite{kang2019few} with more complex feature aggregation module and unify few-shot object detection and viewpoint estimation. 
Another line is fine-tuning based approaches \cite{wang2020few,sun2021fsce,li2021few,fan2021generalized,cao2021few,qiao2021defrcn}.
TFA~\cite{wang2020few} firstly introduces a simple base-training and few-shot fine-tuning paradigm. 
FSCE~\cite{sun2021fsce} improves the TFA baseline by fine-tuning more layers and brings batch contrastive learning to FSOD.
FADI~\cite{cao2021few} divides the fine-tuning stage into association and discrimination to promote the discriminate power of the classifier.
DeFRCN~\cite{qiao2021defrcn} devises GDL and PCB to alleviates the potential contradictions of Faster R-CNN~\cite{ren2015faster} in FSOD.

\subsection{Semi-Supervised Object Detection}
Semi-Supervised Object Detection (SSOD) aims to train a detector with limited labeled data and abundant unlabeled data. 
There are two lines of methods, the consistency methods~\cite{jeong2019consistency,tang2021proposal} and pseudo label methods~\cite{sohn2020detection,tang2021humble,liu2021unbiased,zhou2021instant,xu2021end}.
CSD~\cite{jeong2019consistency} enforces a consistency loss between the original image and the horizontally flipped one.
STAC~\cite{sohn2020detection} proposes a simple pseudo-labeling framework, which trains the model with highly confident pseudo labels from unlabeled dataset with strong augmentations.
Unbiased Teacher~\cite{liu2021unbiased} finds the bias existed in pseudo labels due to over-fitting and class imbalance, hence introducing EMA and Focal Loss~\cite{lin2017focal} to resolve them. There are many subsequent variants~\cite{tang2021humble,zhou2021instant,xu2021end}.
Our work shares similar ideas with pseudo labeling methods, but it is not feasible to directly apply SSOD methods.
Due to the severe data scarcity and extreme class imbalance, the poor-learned teacher model cannot well discover potential novel instances. Moreover, SSOD methods usually rely on a heuristics confidence threshold which fails to implicate the quality of novel instances in FSOD scenario. 
Hence we propose \emph{MINI} to better tackle it.

\subsection{Self-Supervised Learning}
Self-supervised learning (SSL), also named representation learning, aims to learn general visual representation for downstream tasks.
Early works rely on ad-hoc heuristics to design pretext tasks~\cite{doersch2015unsupervised,zhang2016colorful,noroozi2016unsupervised,komodakis2018unsupervised}, which limits the generality of learned representations.
Recent approaches can be categorized as discriminative~\cite{dosovitskiy2014discriminative,bachman2019learning,he2020momentum,chen2020simple,grill2020bootstrap} or generative~\cite{kingma2013auto,xie2021simmim,bao2021beit,he2021masked}.
Contrastive methods~\cite{dosovitskiy2014discriminative,bachman2019learning,he2020momentum,chen2020simple} are representative for discriminative methods, which enforce a consistency loss between different views of the same image by contrastive positive pairs against negative pairs, has shown promising results recently.
We notice the unsupervised learned visual representation by SSL pre-training has strong discriminative power, and we exploit it for better instance mining.

%% file: sections/method.tex
\section{Our Approach}
\label{sec:method}

In this section, we first revisit the problem setting of the conventional few-shot object detection, and discuss limitations of the widely adopted pretrain-transfer paradigm. Then we elaborate our \emph{Mining Implicit Novel Instances (MINI)} to better tackle it.

\begin{figure}[t]
    \centering
    \includegraphics[width=1.\linewidth]{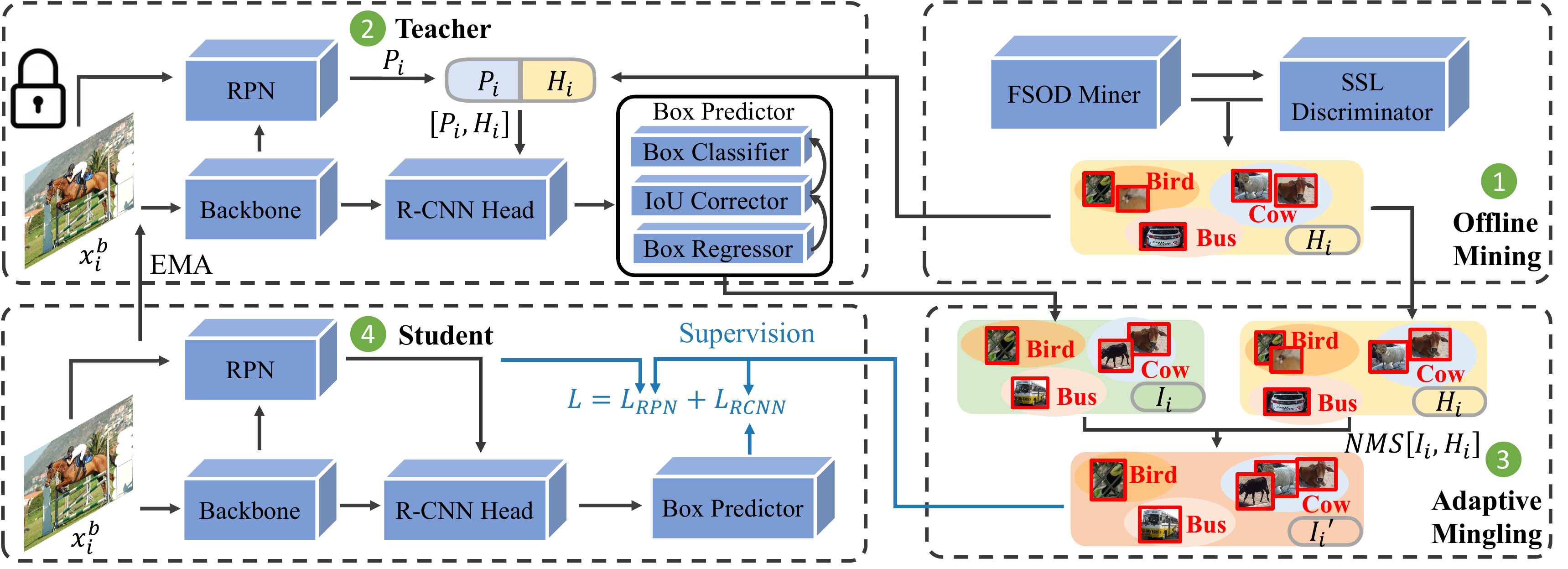}
    \caption{\bd{Method Overview}. 
    \emph{MINI} mines implicit novel instances with an offline mining mechanism and online mining mechanism.
    The pipeline of \emph{MINI} is following:
    1) An FSOD detector is used to discover initial implicit novel instances.
    The offline mining mechanism leverages a self-supervised discriminative model to calibrate classification confidences of these discovered novel instances. 
    2) In the online mining mechanism, the teacher model discovers the implicit novel instances in each iteration during training. 3) The offline and online discovered novel instances are combined with an adaptive mingling design. 4) The student model takes implicit novel instances as ground-truths and updates the parameters of the teacher model via EMA. 
    }
    \label{fig:framework-online}
    \vspace{-6mm}
\end{figure}

\subsection{Revisiting of Few-Shot Object Detection (FSOD)}
\label{sec:define}

In the conventional few-shot object detection (FSOD), there exists two non-overlapping training sets, \ie, a base dataset $D^b=\{x_i^b, y_i^b\}$ with exhaustive annotated instances for each base class $C^b$, and a novel dataset $D^n=\{x_i^n, y_i^n\}$ with $K$ annotated instances for each novel class $C^n$, here $x_i, y_i$ indicates the input image and ground truth, respectively. The ultimate goal of FSOD is to optimize a robust detector to detect objects in a test set that comprises both classes in $C^b\cup C^n$.

To leverage abundant base dataset $D^b$, most FSOD works follow a \emph{pretrain-transfer} paradigm, where the model is firstly pre-trained on $D^b$ to attain general representation ability, and then being transferred to novel class with $K$ few-shot novel samples in $D^n$. 
Due to the limited number of novel samples, most parameters of the FSOD model are frozen when being transferred to preserve the pre-trained knowledge and prevent over-fitting.

Despite the substantial progress in FSOD area, due to the co-occurrence between base and novel classes in $D^b$, the model is learned to treat the co-occurred novel classes as backgrounds.
During transferring, given only $K$ novel samples from $D^n$, such classification bias is hard to be eliminated with frozen most parameters of the pre-trained model.

To overcome the obstacles, we propose \emph{Mining Implicit Novel Instances (MINI)} to mine implicit novel instances with an offline mining mechanism and an online mining mechanism.
As shown in Fig~\ref{fig:framework-online}, we first train an FSOD detector to discover initial implicit novel instances (Sec.~\ref{sec:fsod-ini}).
The offline mining mechanism leverages a self-supervised discriminative model to calibrate classification confidences of these discovered novel instances (Sec.~\ref{sec:offline-mine}). During training, the online mining mechanism takes a teacher-student framework to simultaneously update the FSOD network and the mined implicit novel instances on the fly (Sec.~\ref{sec:online-train}). 
Specifically, the teacher model discovers the implicit novel instances in each iteration during training. The offline and online discovered novel instances are combined with an adaptive mingling design. The student model takes updated implicit novel instances as ground-truths and updates the parameters of the teacher model via Exponential Moving Average (EMA)~\cite{kingma2014adam,he2020momentum,liu2021unbiased}. 

\subsection{FSOD as Initial Miner}
\label{sec:fsod-ini}

In this section, we aim to obtain an object detector that has some basic ability to recognize novel classes.
The initial FSOD network can be readily instantiated with different FSOD algorithms. For simplicity, we adopt the widely used TFA~\cite{wang2020few} in this work, which divides the whole training pipeline into two independent stages as follows,

\paragraph{\bd{Base Model Training Stage}} 
In the first base training stage, the whole model, including the box predictors, \ie, the classifier and regressor, and the feature extractor, \ie, the rest of the network, are jointly trained on the base dataset $D^b$ with abundant annotations of base classes.
To this end, the base model learns a general feature representation ability and is ready to transfer to novel classes.

\paragraph{\bd{Few-Shot Fine-tuning Stage}} 
In the second few-shot fine-tuning stage, only the box predictor is fine-tuned on a small balanced training set that comprises both base and novel classes. The feature extractor will be frozen to preserve the pre-trained general knowledge and prevent the potential over-fitting on the scarce novel set.

\subsection{Offline Mining Mechanism}
\label{sec:offline-mine}

After initializing an FSOD model $M^s$ that can detect novel categories,
in this section, we aim to discover implicit novel instances from $D^b$ with $M^s$ in an offline manner.
Specifically, we preform inference of $M^{s}$ over each image $x_i^b \in D^b$.
The mining process can be formulated as follows,
\begin{equation}
    \label{equ:mining}
    P_i=\phi^{RPN}(x_i^b),
    \ \ \ \ \ \\
    \hat{Y_i} = \phi^{RCNN}(x_i^b, P_i),
    \ \ \ \ \ \\
    h_i = \{\hat{y}_{ij}|s_{ij} >= \delta\}.
\end{equation}
The RPN $\phi^{RPN}$ first predicts a set of proposals $P_i$, the R-CNN $\phi^{RCNN}$ classifies and regress each of proposal $p_{ij} \in P_i$, then some post-processing procedures, \eg, NMS, are applied to yield the inference results $\hat{Y_i}=\{s_{ij}, b_{ij}, l_{ij}\}$, where $s_{ij}, b_{ij}, l_{ij}$ denotes the predicted score, bounding box and label of $j_{th}$ candidate instance on $i_{th}$ image, respectively.
A fixed high confidence threshold $\delta$, \eg, 0.9, is set to filter boxes of low quality in $\hat{Y}_i$.
To this end, the remaining instances $h_i = \{(s_{ij}, b_{ij}, l_{ij})\}$, will be added to the offline novel instances pool $H$.

Although the fixed threshold method receives remarkable success in semi-supervised object detection (SSOD)~\cite{sohn2020detection,liu2021unbiased}, it is not sufficient under the scenario of FSOD.
The severe data scarcity and extreme class imbalance make the predicted novel scores of $M^s$ exhibit a large variance and tend to be generally low, hence the fixed high confidence threshold fails to deal with different novel classes.
On the other hand, the pervasive misclassification of the novel classifier of $M^s$ results in massive false positives in $H$.
Towards the aforementioned drawbacks, we introduce \bd{co-mining with self-supervised discriminator} to promote the discriminative ability of the classifier.
Furthermore, \bd{adaptive thresholding} is proposed to find a proper threshold for different novel classes.

\begin{figure}[t]
    \centering
    \includegraphics[width=1.0\linewidth]{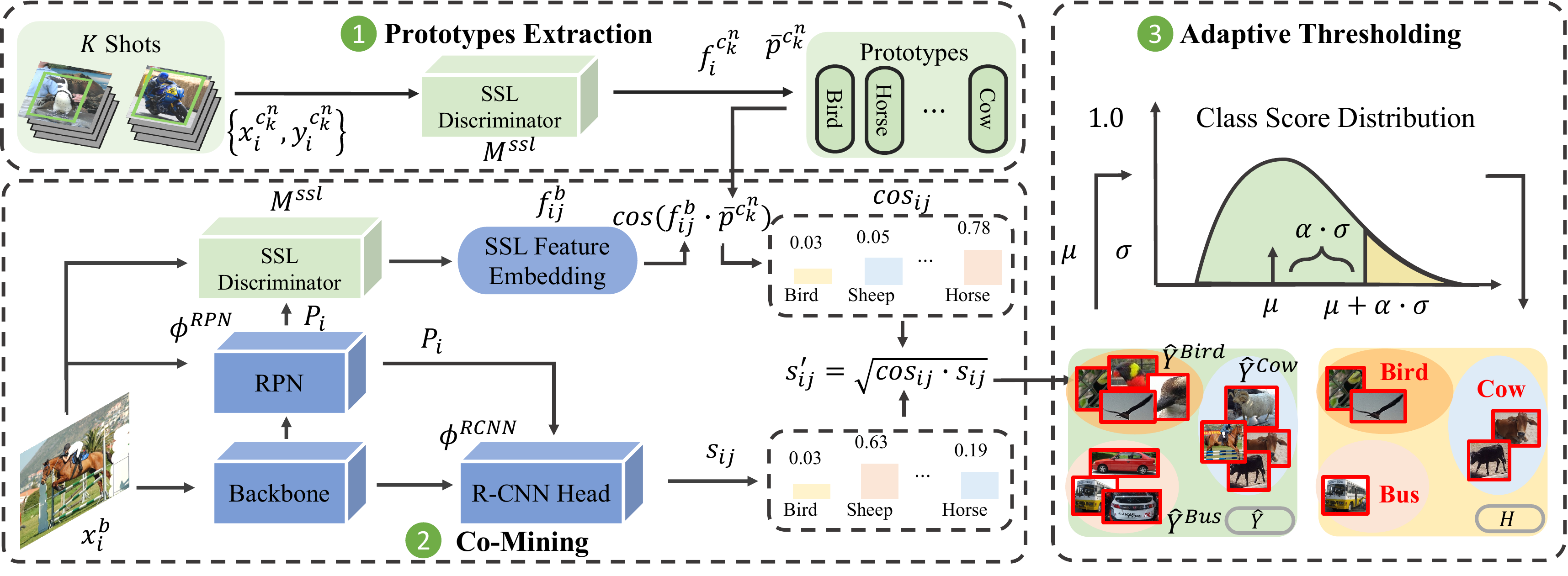}
    \caption{
    Pipeline of offline mining mechanism. SSL model first extracts class prototypes from $D^n$.
    The FSOD detector performs inference on $D^b$, and the SSL model calibrates its scores via calculating cosine similarities between the class prototypes and the box features.
    Adaptive thresholding then computes class-wise statistics from calibrated boxes to determine a proper threshold to filter out mined instances with low quality.
    }
    \label{fig:framework-offline}
    \vspace{-6mm}
\end{figure}

\paragraph{\bd{Co-Mining with Self-Supervised Discriminator}}

Given only the training samples of $K$ shots, it is challenging to acquire a discriminative classifier.
Inspired by the latest development in self-supervised learning (SSL)~\cite{chen2020mocov2}, the self-supervised visual representations are incorporated with strong discriminative power. Hence we propose a novel co-mining scheme that leverages a SSL model $M^{ssl}$ to collaborative mine implicit novel instances with $M^s$.

As shown in Fig.~\ref{fig:framework-offline}, given $K$ instances $\{x_i^{c_k^n}, y_i^{c_k^n}\}$ for each novel class $c_k^n$ from $D^n$, we first forward the image $x_i^{c_k^n}$ through $M^{ssl}$, then employ RoIAlign~\cite{he2017mask} to extract the area bounded by the ground-truth box $b_i^{c_k^n}$ as follows,
\begin{equation}
    \label{equ:ssl-roi-align}
    f_i^{c_k^n} = \text{RoIAlign}(M^{ssl}(x_i^{c_k^n}), b_i^{c_k^n}),
    \ \ \ \ \ \\
    \bar{p}^{c_k^n} = \frac{1}{K} \sum_{i=1}^K f_i^{c_k^n},
\end{equation}
where $f_i^{c_k^n}$ denotes the feature embedding of $i_{th}$ instance, and the class prototype $\bar{p}^{c_k^n}$ is the mean feature over all $K$ instances.
During inference in Equ.~\ref{equ:mining}, the RPN first predicts a set of proposals $P_i$. We then compute the feature embedding $f_{ij}^b$ of $j_{th}$ proposal $p_{ij}\in P_i$ on $i_{th}$ image $x_i^b \in D^b$ similar to Equ.~\ref{equ:ssl-roi-align}, and cosine similarity score is computed with the class prototype $\bar{p}^{c_k^n}$ for each novel class $c_k^n$,
\begin{equation}
    \label{equ:ssl-cosine}
    f_{ij}^b = \text{RoIAlign}(M^{ssl}(x_i^b), p_{ij}),
    \ \ \ \ \ \\
    cos_{ij}^{c_k^n}=\frac{\tau \cdot {f_{ij}^b}^T \bar{p}^{c_k^n}}{||f_{ij}^b|| \cdot ||\bar{p}^{c_k^n}||},
\end{equation}
where $\tau$ is the temperature factor.
We concatenate all cosine similarities of $N$ novel classes and apply the calibration as follows,
\begin{equation}
    \label{equ:ssl-calibrate}
    cos_{ij}=[cos_{ij}^{c_1^n}, \cdots, cos_{ij}^{c_N^n}]
    \ \ \ \ \ \\
    s_{ij}' \leftarrow \sqrt{cos_{ij} \cdot s_{ij}},
\end{equation}
where $[\cdots]$ denotes concatenation operation. It is noted we only apply the calibration on the novel part of $s_{ij}$.
To this end, all the inference results $\hat{Y}_i$ will be collected as $\hat{Y}$ for filtering in the next step.

\paragraph{\bd{Adaptive Thresholding}}
\label{sec:ada-thr}

To filter candidate instances of low quality in $\hat{Y}$, we propose an adaptive thresholding scheme to obtain a proper $\delta$ according to the class-wise score distributions.
As shown in Fig.~\ref{fig:framework-offline}, for each novel classes $c_k^n \in C^n$, we first extract its candidate instances set $\hat{Y}^{c_k^n} = \{s_i^{c_k^n}, b_i^{c_k^n}, l_i^{c_k^n}\}$ from $\hat{Y}$. 
After then, the mean $\mu^{c_k^n}$ and deviation $\sigma^{c_k^n}$ will be computed based on classification scores $\{s_i^{c_k^n}\}$. To this end, we can compute the final confidence threshold and filter low quality predict results as follows,
\begin{equation}
    \label{equ:ada-thr}
    \delta^{c_k}=\mu^{c_k^n} + \alpha \cdot \sigma^{c_k^n},
    \ \ \ \ \ \\
    H^{c_k^n} = \{\hat{y}_i^{c_k^n}|s_{i} >= \delta^{c_k^n}\},
\end{equation}
where $\alpha$ is a coefficient that controls the magnitude of deviation offset to decide the number of kept instances. 
It is noted we further clamp the maximum number to be $N$.
Intuitively, the score mean $\mu^{c_k^n}$ is a measure of the transferring hardness of novel class $c_k^n$, and $\sigma^{c_k^n}$ indicates the compactness of intra-class score distribution. The $\delta^{c_k^n}$ leverages both the $\mu^{c_k^n}$ and $\sigma^{c_k^n}$ to adaptively distinguish the reliable implicit novel instances without introducing computational cost.


\subsection{Online Mining Mechanism}
\label{sec:online-train}

With the offline mined novel instances $H$, we are ready to re-train a new detector with satisfactory performance. 
However, these instances are sourced from a static offline teacher $M^{s}$ of limit precision, and cannot be updated as the model improves, which hinders the further performance improvement.
Hence we introduce an online mining mechanism to update $H$ on the fly.
Specifically, we adopt a teacher-student learning paradigm as shown in Fig.~\ref{fig:framework-online}.
During training, the student is supervised by the mined novel instances. 
The teacher shares the same network architecture with the student model. And its parameters are update by exponential moving average (EMA) of the student's parameters. 
The slowly updated teacher can be considered a temporal model ensemble of the student at different iterations, hence detecting implicit novel instances more accurately.

After mining implicit novel isntance $I$ with the teacher model in each iteration on the fly, the next question is how to update the offline mined novel instances $H$ with $I$.
It is noteworthy that the poor-learned teacher fail to discover valuable novel instances at the beginning of the training.
Thus, we devise a concise adaptive mingling scheme, where the offline and online mined instances are adaptive balanced as the training process proceeds.
We further introduce an IoU branching mechanism to improve the quality of online mined novel instances.

\paragraph{\bd{Adaptive Mingling}}

During training, given a training sample $x_i^b$, the teacher first online mines novel instances $I_i$ with a similar procedure with Equ.~\ref{equ:mining}, and we mingle the online mined novel instances $I_i$ with the offline mined novel instances $H_i$ as follows,
\begin{equation}
    \label{equ:ada-bal}
    \hat{Y}_i = \phi^{RCNN}(x_i^b, [P_i, H_i]),
    \ \ \ \ \ \\
    I_i = \{\hat{y}_{ij}|s_{ij} >= \delta\}, 
    \ \ \ \ \ \\
    I_i' = NMS([I_i, H_i]).
\end{equation}
We concatenate $H_i$ with $P_i$ and $I_i$, respectively. Here $P_i$ is RPN proposals predicted by the teacher model. We argue these two concatenations play an important role from two aspects.
1) At the beginning of the training, due to the poor-learned RPN and R-CNN, the high confidence threshold $\delta$ can filter almost all of the novel instances, so that $I_i$ degrades to an empty set $\phi$, hence only $H_i$ is remained to provide training signal to warm up the beginning training of the student.
2) As the training process proceeds, the online teacher becomes more and more discriminative. By presenting $H_i$ as extra proposals, the teacher will calibrate some misclassification in $H_i$. Moreover, the teacher can also discover missed instances in $H_i$.
The mingled instances $I_i'$ work as ground-truths of novel classes during the training of the student model.

\paragraph{\bd{IoU Branching Correction}}

To further improve the quality of online mined novel instances, we notice the model trained under low data regime cannot well recognize precisely-localized boxes, hence we introduce IoU Branching mechanism to better mine high quality novel instances.
Specifically, we introduce an extra IoU branch that parallels to the original R-CNN head to learn to predict the IoU between predicted boxes and ground truths. The structure is the same as the original R-CNN branch, \ie, two fully-connected (FC) layers and followed by an IoU predictor (a single FC layer). 
During mining, we combines the classification scores with IoU scores in Equ.~\ref{equ:mining} as follows,
\begin{equation}
    \label{equ:iou-mining}
    s_{ij}' = \sqrt{s_{ij} \cdot iou_{ij}},
    \ \ \ \ \ \\
    I_i = \{\hat{y}_{ij}|s_{ij}' >= \delta\}, 
\end{equation}
where $iou_{ij}$ denotes the predicted IoU score of $j_{th}$ proposal on $i_{th}$ image.
And 
A standard MSE loss is adopted to optimize the IoU branch.
All modules of the R-CNN head are jointly optimized by the following loss in an end-to-end manner:
\begin{equation}
    \label{equ:loss}
    \mathcal{L}_{RCNN}=\mathcal{L}_{cls} + \mathcal{L}_{reg} + \beta \cdot \mathcal{L}_{IoU},
\end{equation}
where $\beta$ denotes the loss weight of the loss of the IoU branch.

%% file: sections/experiments.tex
\section{Experiments}
\label{sec:experiments}

In this section, we first outline the datasets and benchmark protocols in Sec~\ref{sec:setting}, the implementation details of our method in Sec~\ref{sec:imp-detail}.
Then, we compare our approach with the latest methods of FSOD and SSOD in Sec~\ref{sec:main-results}.
Finally, we make an extensive ablation study about different components in Sec~\ref{sec:aba}.

\subsection{Datasets and Evaluation Protocols}
\label{sec:setting}

We follow the same data split construction and evaluation protocols used in ~\cite{wang2020few} for fair comparisons. 
All experiments are evaluated on both PASCAL VOC~\cite{Everingham10} and MS COCO~\cite{lin2014microsoft} datasets.

\paragraph{\bd{PASCAL VOC}} has 20 classes, which are randomly split into 15 base classes and 5 novel classes. There are 3 different class splits, and we refer them as Novel Split 1, 2 and 3, respectively.
For each split, there exists exhaustive base instances but only $K=1, 2, 3, 5, 10$ annotated instances for novel classes.
All instances are sampled from the union of VOC07 and VOC12 train/val set for training, and the model is tested on VOC07 test set.
The standard PASCAL VOC metric, \ie, Average Precision (IoU=0.5) for novel classes (nAP50) is reported.

\paragraph{\bd{MS COCO}} has 80 classes, 20 classes that overlap with PASCAL VOC are regarded as novel classes, the remaining 60 classes are considered as base classes.
We evaluate our method for $K=10, 30$ shots. 
And the standard COCO-style metric is adopted, which averages mAP of IoUs from 0.5 to 0.95 with an interval of 0.05. We also report nAP50 and nAP75, respectively.

\subsection{Implementation Details}
\label{sec:imp-detail}

We implement our method based on MMDetection~\cite{mmdetection} and MMFewShot~\cite{mmfewshot2021}.
We employ the Faster R-CNN~\cite{ren2015faster} with Feature Pyramid Network~\cite{lin2017feature} and ResNet-101~\cite{he2016deep} as base model. Please refer to the Appendix for the detailed settings.

\subsection{Main Results}
\label{sec:main-results}

\begin{table*}[t]
    \caption{\small{Performance (novel AP50) across three splits on PASCAL VOC dataset. \bd{\R{Red}}/\bd{\B{Blue}} denote best and second-best results, respectively}}
    \vspace{2mm}
    \label{tab:voc_main}
    \centering
    \begin{adjustbox}{width=\textwidth}
    \setlength\tabcolsep{3.5pt}
    \begin{tabular}{@{}l|ccccc|ccccc|ccccc@{}}
        \toprule
        \multirow{2}{*}{Method/Shot}                                  & \multicolumn{5}{c|}{Novel Split 1}                                            & \multicolumn{5}{c|}{Novel Split 2}                                            & \multicolumn{5}{c}{Novel Split 3}                                                  \\
                                                                      & 1             & 2             & 3             & 5             & 10            & 1             & 2             & 3             & 5             & 10            & 1             & 2             & 3             & 5             & 10            \\
        \midrule
        FSRW   ~\cite{kang2019few}        \hfilll \textit{ICCV 19}    & 14.8          & 15.5          & 26.7          & 33.9          & 47.2          & 15.7          & 15.3          & 22.7          & 30.1          & 40.5          & 21.3          & 25.6          & 28.4          & 42.8          & 45.9          \\ 
        MetaDet~\cite{wang2019meta}       \hfilll \textit{ICCV 19}    & 18.9          & 20.6          & 30.2          & 36.8          & 49.6          & 21.8          & 23.1          & 27.8          & 31.7          & 43.0          & 20.6          & 23.9          & 29.4          & 43.9          & 44.1          \\
        Meta R-CNN~\cite{yan2019meta}     \hfilll \textit{ICCV 19}    & 19.9          & 25.5          & 35.0          & 45.7          & 51.5          & 10.4          & 19.4          & 29.6          & 34.8          & 45.4          & 14.3          & 18.2          & 27.5          & 41.2          & 48.1          \\
        TFA w/ cos~\cite{wang2020few}     \hfilll \textit{ICML 20}    & 39.8          & 36.1          & 44.7          & 55.7          & 56.0          & 23.5          & 26.9          & 34.1          & 35.1          & 39.1          & 30.8          & 34.8          & 42.8          & 49.5          & 49.8          \\
        MPSR~\cite{wu2020multi}           \hfilll \textit{ECCV 20}    & 41.7          & -             & 51.4          & 55.2          & 61.8          & 24.4          & -             & 39.2          & 39.9          & 47.8          & 35.6          & -             & 42.3          & 48.0          & 49.7          \\
        FSCE~\cite{sun2021fsce}           \hfilll \textit{CVPR 21}    & 44.2          & 43.8          & 51.4          & 61.9          & \bd{\B{63.4}} & 27.3          & 29.5          & 43.5          & 44.2          & \bd{\B{50.2}} & 37.2          & 41.9          & 47.5          & 54.6          & 58.5          \\
        SRR-FSD~\cite{zhu2021semantic}    \hfilll \textit{CVPR 21}    & 47.8          & 50.5          & 51.3          & 55.2          & 56.8          & \bd{\B{32.5}} & 35.3          & 39.1          & 40.8          & 43.8          & 40.1          & 41.5          & 44.3          & 46.9          & 46.4          \\
        CME~\cite{li2021beyond}           \hfilll \textit{CVPR 21}    & 41.5          & 47.5          & 50.4          & 58.2          & 60.9          & 27.2          & 30.2          & 41.4          & 42.5          & 46.8          & 34.3          & 39.6          & 45.1          & 48.3          & 51.5          \\
        TIP~\cite{li2021transformation}   \hfilll \textit{CVPR 21}    & 27.7          & 36.5          & 43.3          & 50.2          & 59.6          & 22.7          & 30.1          & 33.8          & 40.9          & 46.9          & 21.7          & 30.6          & 38.1          & 44.5          & 50.9         \\
        FADI~\cite{cao2021few}            \hfilll \textit{NeurIPS 21} & 50.3          & 54.8          & 54.2          & 59.3          & 63.2          & 30.6          & 35.0          & 40.3          & 42.8          & 48.0          & 45.7          & 49.7          & 49.1          & \bd{\B{55.0}} & \bd{\B{59.6}} \\
        DeFRCN~\cite{qiao2021defrcn}      \hfilll \textit{ICCV 21}    & \bd{\B{53.6}} & \bd{\B{57.5}} & \bd{\B{61.5}} & \bd{\B{64.1}} & 60.8          & 30.1          & \bd{\B{38.1}} & \bd{\B{47.0}} & \bd{\B{53.3}} & 47.9          & \bd{\B{48.4}} & \bd{\B{50.9}} & \bd{\B{52.3}} & 54.9          & 57.4          \\
        \bd{MINI (Ours)}                                                     & \bd{\R{72.0}} & \bd{\R{74.2}} & \bd{\R{72.4}} & \bd{\R{74.7}} & \bd{\R{76.2}} & \bd{\R{51.8}} & \bd{\R{53.6}} & \bd{\R{62.3}} & \bd{\R{62.1}} & \bd{\R{63.7}} & \bd{\R{65.0}} & \bd{\R{66.5}} & \bd{\R{64.0}} & \bd{\R{66.9}} & \bd{\R{70.4}} \\
        \bottomrule
    \end{tabular}
    \end{adjustbox}
    \vspace{-5mm}
\end{table*}

\begin{table}[t]
    \caption{\small{Performance on MS COCO dataset. \bd{\R{Red}}/\bd{\B{Blue}} denote best and second-best results, respectively}}
    \vspace{2mm}
    \label{tab:coco_main}
    \centering
    \begin{adjustbox}{width=\textwidth}
    \setlength\tabcolsep{3.5pt}
    \begin{tabular}{@{}l|cc|cc|cc|l|cc|cc|cc@{}}
        \toprule
        \multirow{2}{*}{Method}          & \multicolumn{2}{c|}{nAP}      & \multicolumn{2}{c|}{nAP50}      & \multicolumn{2}{c|}{nAP75}      & \multirow{2}{*}{Method}          & \multicolumn{2}{c|}{nAP}      & \multicolumn{2}{c|}{nAP50}      & \multicolumn{2}{c}{nAP75}      \\
                                         & 10            & 30            & 10             & 30             & 10             & 30             &                                  & 10            & 30            & 10             & 30             & 10             & 30            \\ \toprule
        FSRW~\cite{kang2019few}          & 5.6           & 9.1           & 12.3           & 19.0           & 4.6            & 7.6            & SRR-FSD~\cite{zhu2021semantic}   & 11.3          & 14.7          & 23.0           & 29.2           & 9.8            & 13.5          \\ 
        MetaDet~\cite{wang2019meta}      & 7.1           & 11.3          & 14.6           & 21.7           & 6.1            & 8.1            & CME~\cite{li2021beyond}          & 15.1          & 16.9          & 24.6           & 28.0           & \bd{\B{16.4}}  & \bd{\B{17.8}} \\
        Meta R-CNN~\cite{yan2019meta}    & 8.7           & 12.4          & 19.1           & 25.3           & 6.6            & 10.8           & TIP~\cite{li2021transformation}  & 16.3          & 18.3          & \bd{\B{33.2}}  & \bd{\B{35.9}}  & 14.1           & 16.9          \\
        TFA w/ cos~\cite{wang2020few}    & 10.0          & 13.7          & 19.1           & 24.9           & 9.3            & 13.4           & FADI~\cite{cao2021few}           & 12.2          & 16.1          & 22.7           & 29.1           & 11.9           & 15.8          \\
        MPSR~\cite{wu2020multi}          & 9.8           & 14.1          & 17.9           & 25.4           & 9.7            & 14.2           & DeFRCN~\cite{qiao2021defrcn}     & \bd{\B{18.5}} & \bd{\B{22.6}} & -              & -              & -              & -             \\
        FSCE~\cite{sun2021fsce}          & 11.9          & 16.4          & -              & -              & 10.5           & 16.2           & \bd{MINI (Ours)}                        & \bd{\R{21.8}} & \bd{\R{27.3}} & \bd{\R{38.0}}  & \bd{\R{44.9}}  & \bd{\R{21.5}}  & \bd{\R{28.5}} \\
    \end{tabular}
    \end{adjustbox}
    \vspace{-3mm}
\end{table}

\begin{table}[t]
    \caption{Performance comparison with SSOD methods on PASCAL VOC dataset}
    \label{tab:compare-ssod}
    \vspace{2mm}
    \centering
    \begin{adjustbox}{width=\textwidth}
    \setlength\tabcolsep{3.5pt}
    \begin{tabular}{@{}l|ccccc|l|ccccc@{}}
        \toprule
        \multirow{2}{*}{Method}    & \multicolumn{5}{c|}{nAP50}                                & \multirow{2}{*}{Method}     & \multicolumn{5}{c}{nAP50}                                 \\
                                   & 1         & 2         & 3         & 5         & 10        &                             & 1         & 2         & 3         & 5         & 10        \\ \midrule
         STAC, $\delta=0.5$  \ \   & 38.8      & 59.2      & 60.2      & 64.8      & 66.4      & UB-T, $\delta=0.5$ \ \      & 3.9       & 13.2      & 15.1      & 11.7      & 6.2       \\
         STAC, $\delta=0.7$  \ \   & 19.3      & 48.5      & 57.1      & 65.2      & 66.9      & UB-T, $\delta=0.7$ \ \      & 1.8       & 3.6       & 22.3      & 52.0      & 60.6      \\
         STAC, $\delta=0.9$  \ \   & 0.0       & 7.1       & 16.6      & 22.6      & 57.5      & UB-T, $\delta=0.9$ \ \      & 0.0       & 3.6       & 8.9       & 25.4      & 47.6      \\
         Ours                      & \bd{72.0} & \bd{74.2} & \bd{72.4} & \bd{74.7} & \bd{76.2} & Ours                        & \bd{72.0} & \bd{74.2} & \bd{72.4} & \bd{74.7} & \bd{76.2} \\
        \bottomrule
    \end{tabular}
    \end{adjustbox}
    \vspace{-3mm}
\end{table}

\paragraph{\bd{Comparison with FSOD Methods}}

Table~\ref{tab:voc_main} presents performance comparisons between our method and the latest FSOD methods across three novel splits on the PASCAL VOC benchmark.  
In all splits and shots, \emph{MINI} achieves new SOTA performance and outperforms the second-best by a large margin.
Specifically, \emph{MINI} boosts the current SOTA by 18.4, 16.7, 10.9, 10.6, 12.8 and 19.3, 15.5, 15.3, 8.8, 13.5 and 16.6, 15.6, 11.7, 11.9, 10.8 for $K$=1, 2, 3, 5, 10 on novel split 1, 2 and 3, respectively. 
The significant performance improvements are consistent across shots and splits, but announces more on low-shot scenarios, since in low-shot scenarios the data scarcity is more severe and the mined implicit shots are good alleviation of that.
Similar performance gains can be observed on the MS COCO benchmark. As shown in Table~\ref{tab:coco_main}, \emph{MINI} outperforms all FSOD methods by a large margin with the COCO-style AP metric. Concretely, our method achieves 21.8 and 27.3 and boosts the SOTA performance by 3.3 and 4.7 for $K=$ 10 and 30, respectively. 
The superior performance on both datasets suggests \emph{MINI} can generalize well under different datasets.

\paragraph{\bd{Comparison with SSOD Methods}}
\label{sec:compare-ssod}
In this section, we explore whether it is feasible to directly apply methods from semi-supervised object detection (SSOD) under the scenario of FSOD. We compare \emph{MINI} with two widely used frameworks, STAC~\cite{sohn2020detection} and Unbiased Teacher (UB-T)~\cite{liu2021unbiased} as representatives for offline and online paradigms, respectively.
As shown in Table~\ref{tab:compare-ssod}, the performance of both STAC and unbiased teacher are far behind \emph{MINI}.
We adopt the same hyper-parameters setting with the official paper except for the confidence threshold $\delta$.
The original STAC adopts $\delta=0.9$. We notice such a high threshold can filter all novel instances, decreasing $\delta$ from 0.9 to 0.5 can significantly boost performance in lower shots, \eg, 0.0, 7.1 to 38.8, 59.2 for $K=$ 1 and 2, respectively. But it can harm the performance in higher shots, \eg, nAP50 drops 0.4 and 0.5 when decreasing $\delta$ from 0.7 to 0.5 for $K=$ 5 and 10, respectively, since it will result in more false positives.
For unbiased teacher, we initialize both teacher and student with TFA~\cite{wang2020few} in the burn in stage~\cite{liu2021unbiased}. 
Though unbiased teacher adopts Focal Loss~\cite{lin2017focal}, we notice it is not sufficient to resolve the severe data scarcity and extreme class imbalance in FSOD.
The proposed \emph{MINI} significantly outperforms these SSOD methods, demonstrating the superiority of our method.

\subsection{Ablation Study}
\label{sec:aba}

In this section, we conduct thorough ablation studies on each component of our approach.
We first demonstrate each component can contribute to the overall performance,
then we analyze the effect of different hyper-parameters.
Finally, we explore how and why each component works.
Unless otherwise specified, all experiments are conducted on novel split 1 of PASCAL VOC benchmark.

\begin{table}[t]
    \caption{Effectiveness of each components}
    \label{tab:component}
    \vspace{2mm}
    \centering
    \begin{tabular}{l|ccccc}
        \toprule
        \multirow{2}{*}{Method} & \multicolumn{5}{c}{nAP50}                               \\
                                & 1         & 2         & 3         & 5         & 10        \\ \midrule
         TFA Base               & 41.9      & 49.1      & 49.9      & 58.0      & 58.4      \\
        +TFA Mining             & 19.3      & 48.5      & 57.1      & 65.2      & 66.9      \\
        +Adaptive Thresholding           & 58.1      & 63.3      & 62.5      & 67.7      & 67.5      \\
        +SSL Co-Mining          & 63.5      & 67.7      & 66.8      & 70.3      & 68.8      \\
        +Adaptive Mingling \ \ & 68.3      & 70.0      & 68.7      & 71.3      & 70.8      \\
        +IoU Branching          & 69.9      & 72.5      & 71.7      & 72.7      & 73.8      \\
        +Fine-Tuning            & \bd{72.0} & \bd{74.2} & \bd{72.4} & \bd{74.7} & \bd{76.2} \\
        \bottomrule
    \end{tabular}
    \vspace{-5mm}
\end{table}

\paragraph{\bd{Component Analysis}}

Table~\ref{tab:component} shows the overall performance contribution of each component.
The first row indicates our re-implemented TFA baseline, the performances of all shots are higher than the original implementation~\cite{wang2020few}.
Directly applying TFA to offline mine implicit instances and re-train a detector with these instances leads to limited performance gains on higher shots but worse performance on lower shots.
Our adaptive thresholding rescues the performance degradation and also improves the performance on all shots, which suggests it is vital to set a proper threshold.
SSL Co-Mining results in decent gains in lower shots but lower gain in higher shots, \eg, +5.4 and +1.3 for $K=$ 1 and 10, which demonstrates SSL is a good enhancement to TFA in low shot, but TFA trained with higher shots has a similar discriminative power will SSL model.
The online mining mechanism employs a teacher to mine diverse novel instances and combine it with the offline mined novel instances with an adaptive mingling design, better training samples lead to decent gains in all shots.
The IoU Branching mechanism is orthogonal to all other modules, further improving the performance.
Finally, we fine-tune the re-trained model on the novel set $D^n$ to mitigate the side effect of the inaccurate supervision from mined implicit instances, especially the box error is harmful to the regressor.
Compare with the TFA baseline, our method leads to total +30.1, +25.1, +22.5, +16.7, +17.8 gains for $K=$ 1, 2, 3, 5 and 10, respectively.

\paragraph{\bd{Ablation Study for Hyper-parameters}}

4 hyper-parameters are introduced, $\alpha$ and $N$ for adaptive thresholding, $\delta$ for online mining, and $\beta$ for IoU branching.
The detailed hyper-parameter study is described in the supplementary materials.


\paragraph{\bd{Flexibility of Adaptive Thresholding}}

\begin{figure}[t]
    \begin{minipage}[t]{.45\linewidth}
    \centering
    \captionsetup{width=1.\linewidth}
    \includegraphics[width=1.\linewidth]{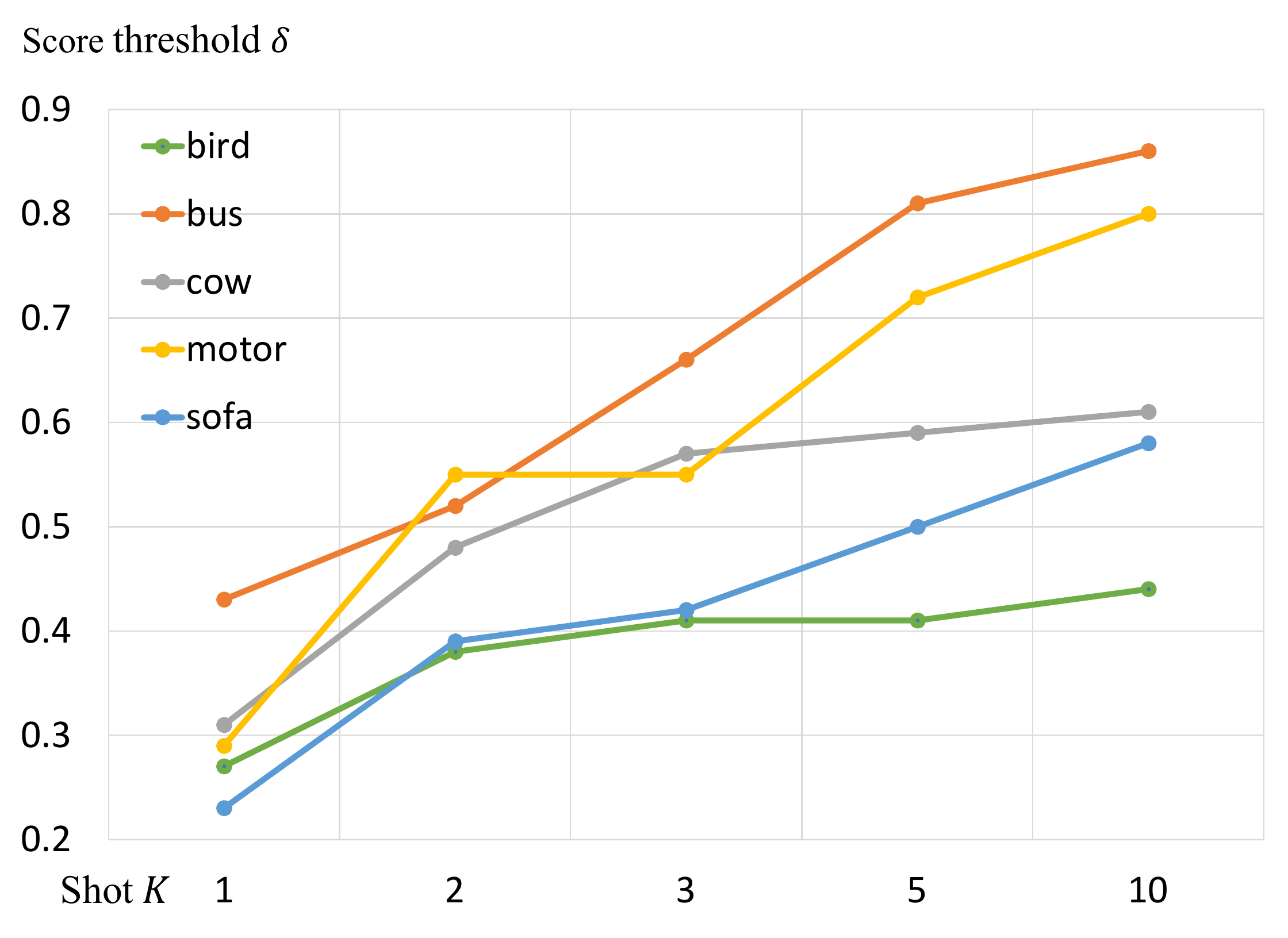}
    \caption{
        Confidence threshold $\delta$ of adaptive thresholding varies among different shot $K$ and novel classes on PASCAL VOC Novel Split 1 
    }
    \label{fig:thr-vs-shot}
    \end{minipage}\hfilll
    \begin{minipage}[t]{.45\linewidth}
    \centering
    \captionsetup{width=1.\linewidth}
    \includegraphics[width=1.\linewidth]{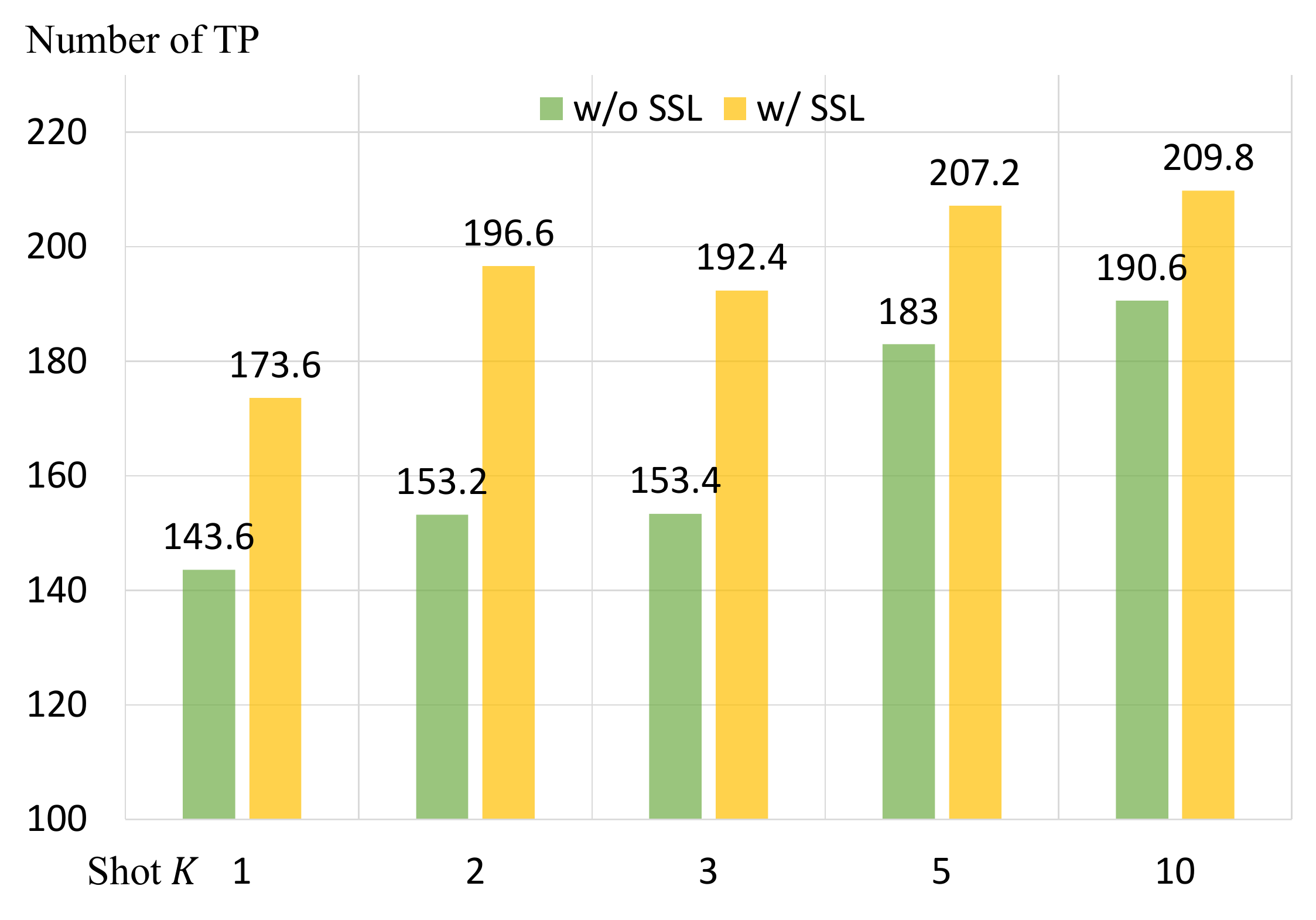}
    \caption{
        Comparison of the number of true positive (TP) of mined novel instances whether applying the SSL co-mining
    }
    \label{fig:ssl-tp-num}
    \end{minipage}
    \vspace{-5mm}
\end{figure}

To understand how adaptive thresholding works, we study how threshold $\delta$ varies among different shots $K$ and classes in Fig.~\ref{fig:thr-vs-shot}.
We can see $\delta$ well characterize the transfer hardness among different classes and shots.
On the one hand, as shot grows, the classification scores should be higher since the classifier learns better.  
Adaptive thresholding decides to steadily increase $\delta$ to rigorous the mined novel instances to suppress false positives. 
On the other hand, the classifier tends to predict higher scores for those novel classes that are similar to base classes~\cite{cao2021few}, \eg, ``bus'' is an easy class since it is similar to ``car'', but ``bird'' is a hard class since no base class is similar to it.
Therefore, adaptive thresholding decides a higher $\delta$ for ``bus'' and a lower $\delta$ for ``bird''.
Such flexibility leads to the strong robustness of our adaptive thresholding to fit in different scenarios.

\paragraph{\bd{Unsupervised SSL Pre-training is Powerful Discriminator}}
Fig~\ref{fig:ssl-tp-num} shows the comparison of the number of true positive (TP) among offline mined novel instances whether applying the SSL co-mining. 
Positive instances are those overlap a GT bounding box with IoU $>=0.5$.
We can see SSL co-mining significantly boosts the number of TP in any shot, especially in low-shot scenarios, \eg, +30.0 and +43.4 in 1 and 2 shots, but the number of increases becomes less as the shot grows.
This aligns with the observations in Tab.~\ref{tab:component} that SSL co-mining brings more gains in low shot.
The FSOD miner learned on scarce novel samples can discover limited implicit novel instances, and the SSL model can greatly enrich the diversity with these extra true positives.

\paragraph{\bd{Complementivity between Offline and Online Mining}}

To understand how adaptive mingling balances online and offline mined instances, we record the number of these two types of instances kept after the NMS in Equ.~\ref{equ:ada-bal} at different iterations in Fig.~\ref{fig:ada-mingling}.
At the beginning of the training, the online teacher mines no instances and offline instances are mainly kept for training.
This well explains the first and second rows of Tab.~\ref{tab:mingling} that it is necessary to enhance the R-CNN, the online teacher cannot discover enough novel instances at the beginning.
As the training process proceeds, online instances gradually dominate kept instances, which demonstrates a better online teacher can discover more diverse novel instances than the initial FSOD detector.
The last row of Tab.~\ref{tab:mingling} shows enhancing the RPN can also bring a slight gain.

\paragraph{\bd{Generalizing to External Datasets}}
Currently, we only mine implicit novel instances from the base dataset, can we generalize \emph{MINI} to external unlabeled

\begin{table}[t]
    \caption{Generalizing \emph{MINI} to mine novel instances from other unlabeled datasets }
    \label{tab:generalization}
    \vspace{2mm}
    \centering
    \subfloat[\small PASCAL VOC \label{tab:generalization-voc}]{
    \begin{tabular}{cc|ccccc}
        \toprule
         Base Set              & Extra Set              & \multicolumn{5}{c}{nAP50}                                 \\ \midrule
         PASCAL VOC            & COCO                   & 1         & 2         & 3         & 5         & 10         \\ \midrule
         \xmark                & \xmark                 & 41.9      & 49.1      & 49.9      & 58.0      & 58.4       \\
         \cmark                & \xmark                 & 72.0      & 74.2      & 72.4      & 74.7      & 76.2       \\
         \xmark                & \cmark                 & 62.9      & 69.3      & 68.2      & 72.9      & 72.4       \\
         \cmark                & \cmark                 & \bf{73.7} & \bf{75.9} & \bf{76.5} & \bf{78.1} & \bf{77.1}  \\ 
        \bottomrule
    \end{tabular}}\hfill
    \subfloat[\small MS COCO \label{tab:generalization-coco}]{
    \begin{tabular}{cc|cc}
        \toprule
        Base Set              & Extra Set              & \multicolumn{2}{c}{nAP} \\ \midrule
        COCO                  & Object365              & 10        & 30          \\ \midrule
        \xmark                & \xmark                 & 10.4      & 14.7        \\
        \cmark                & \xmark                 & 21.8      & 27.3        \\
        \xmark                & \cmark                 & 21.2      & 26.4        \\
        \cmark                & \cmark                 & \bf{23.6} & \bf{29.3}   \\ 
        \bottomrule
    \end{tabular}}
    \vspace{-8mm}
\end{table}

\begin{minipage}[t]{1.\linewidth}
    \vspace{-15mm}
    \begin{minipage}[t]{.4\linewidth}
        \vspace{-33mm}
        \centering
        \includegraphics[width=1.\linewidth]{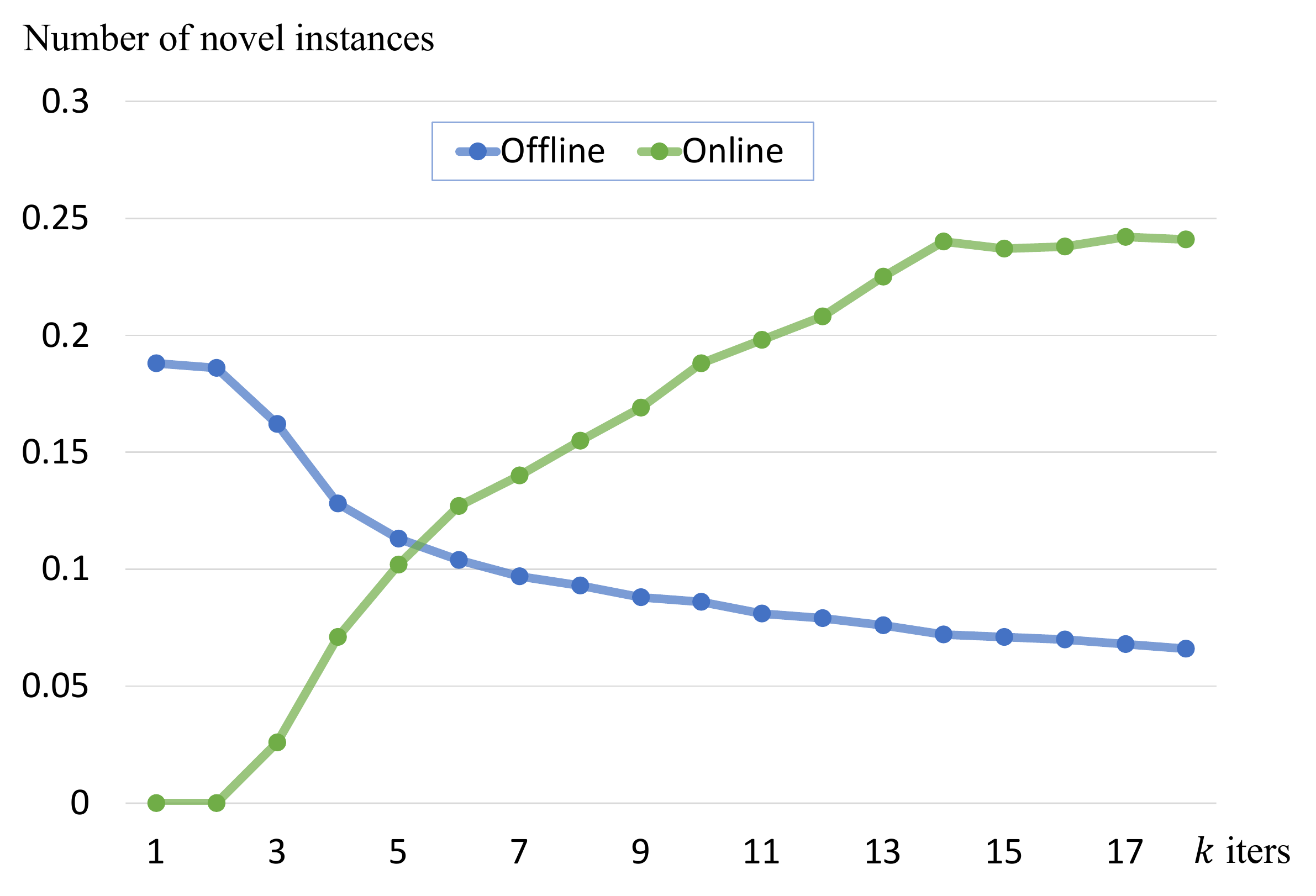}
        \captionof{figure}{Number of offline and online mined instances kept for training at different iterations}
        \label{fig:ada-mingling}
    \end{minipage}\hspace{10mm}
    \begin{minipage}[b]{.45\linewidth}
        \vspace{12mm}
        \centering
        \captionof{table}{Ablation study for whether enhancing RPN or R-CNN of the online teacher with offline mined novel instances}
        \label{tab:mingling}
        \begin{adjustbox}{width=1.\textwidth}
        \setlength\tabcolsep{2pt}
        \begin{tabular}{cc|ccccc}
            \toprule
            \multirow{2}{*}{RPN} & \multirow{2}{*}{R-CNN} & \multicolumn{5}{c}{nAP50}                               \\
                                 &                        & 1         & 2         & 3         & 5         & 10        \\ \midrule
            \xmark               & \xmark                 & 0.0       & 0.0       & 0.0       & 0.0       & 0.0       \\
            \cmark               & \xmark                 & 0.0       & 0.0       & 0.0       & 0.0       & 0.0       \\
            \xmark               & \cmark                 & 69.5      & 72.1      & 70.9      & 71.5      & 72.6      \\
            \cmark               & \cmark                 & \bd{69.9} & \bd{72.5} & \bd{71.7} & \bd{72.7} & \bd{73.8} \\
            \bottomrule
        \end{tabular}
        \end{adjustbox}
    \end{minipage}
    \vspace{4mm}
\end{minipage}

\noindent datasets in a cross-domain manner?
In this section, we explore two such settings, which adopt MS COCO~\cite{lin2014microsoft} and Object365~\cite{shao2019objects365} as external datasets for the original base set PASCAL VOC~\cite{Everingham10} and MS COCO, respectively.
The results are shown in Tab.~\ref{tab:generalization}.
We adopt same hyper-parameters and mine 100 and 2000 instances on extra datasets for each novel class in Tab.~\ref{tab:generalization-voc} and Tab.~\ref{tab:generalization-coco}, respectively.
Mining only from base or extra set can both significantly improve the performance, but the performance of extra set is inferior to base set due to the domain gap between datasets.
Moreover, mining from both sets can further bring considerable gains, which demonstrates \emph{MINI} can well generalize to external datasets and discover valuable instances to enhance the original model.



%% file: sections/conclusion.tex
\section{Conclusion}
\label{sec:conclusion}

In this paper, we propose \emph{Mining Implicit Novel Instances (MINI)} to better tackle FSOD.
MINI comprises an offline mining mechanism and an online mining mechanism.
The offline mining mechanism leverages a self-supervised discriminator to collaboratively mine implicit novel instances with a trained FSOD model.
Taking the mined novel instances as auxiliary training samples, the online mining mechanism takes a teacher-student framework to simultaneously update the FSOD model and the mined implicit novel instances on the fly.
\emph{MINI} achieves new SOTA performance on various benchmarks, which demonstrates its effectiveness.

%% file: sections/appendix.tex
\section*{Appendix A: Implementation Details}
We implement our method based on MMDetection~\cite{mmdetection} and MMFewShot~\cite{mmfewshot2021}.
We employ the Faster R-CNN~\cite{ren2015faster} with Feature Pyramid Network~\cite{lin2017feature} and ResNet-101~\cite{he2016deep} as base model. 
All models are trained on 8 Titan-XP GPUs with batch-size 16 (2 images per GPU), and optimized by a standard SGD optimizer with learning rate 0.02, momentum 0.9 and weight decay $10e^{-4}$.
We strictly follow the protocol introduced by TFA~\cite{wang2020few} without any modifications to initialize the $\cM^s$.
MoCo v2~\cite{chen2020mocov2} w/ ResNet-50~\cite{he2016deep} is employed to co-mine novel instances with FSOD model, and we take the C4 feature, \ie, the feature of the last layer of ResNet to compute the cosine similarity. $\alpha$ in Equ. 5 is set to be 1.5 for all experiments, and we limit the maximum of novel instances to be 300 and 3000 for PASCAL VOC and MS COCO, respectively.
For the online learning stage, we follow unbiased teacher~\cite{liu2016large} to apply weak and strong augmentations to the teacher and student model, respectively. 
For PASCAL VOC all models are trained for 18k iterations and decayed at 12k and 16k, respectively, and the confidence threshold $\delta$ is set to be 0.7.
For MS COCO all models are trained for 160k iterations and decayed at 110k and 145k, respectively, and the confidence threshold $\delta$ is set to be 0.8.
In the last fine-tuning stage, for PASCAL VOC, we only fine-tune the box classifier, predictor and IoU predictor for 4k, 8k, 8k, 8k, 12k iterations for $K=1, 2, 3, 5, 10$, respectively.
For MS COCO, we fine-tune the whole R-CNN head for 4k, 8k iterations for $K=10, 30$, respectively. The learning rate is set as 0.001 for both datasets.

\begin{table}[t]
    \centering
    \caption{
        Performance comparison between TFA and \emph{MINI}.
        We use the same base and novel set as PASCAL VOC novel split1 with the shot1 setting, but we exclude images that contain a selected novel class $c^n$ from the base dataset to simulate there is no co-occurred $c^n$, and the excluded class is marked as \R{red}.
        For example, the ``bird'' row indicates we exclude all the images contain ``bird'' instances from the base dataset
    }
    \label{tab:robustness}
    \vspace{2mm}
    \subfloat[\small TFA \label{tab:robustness-tfa}]{
    \begin{tabular}{l|c|ccccc}
       \toprule 
       Exclude   & nAP & Bird & Bus & Cow & Motor & Sofa \\ 
       \midrule
       Bird      & 37.2 & {\R{27.1}}    & 63.8          & 33.0          & 45.0          & 17.2 \\
       Bus       & 37.9 & 24.0          & {\R{39.4}}    & 39.0          & 59.7          & 27.5 \\
       Cow       & 38.5 & 22.6          & 54.9          & {\R{43.0}}    & 54.9          & 16.9 \\
       Motor     & 36.6 & 18.4          & 57.7          & 41.2          & {\R{42.7}}    & 23.0 \\
       Sofa      & 41.8 & 27.3          & 70.0          & 37.1          & 49.3          & {\R{25.1}} \\
       \bottomrule
    \end{tabular}}\hfill
    \subfloat[\small MINI \label{tab:robustness-mini}]{
    \begin{tabular}{l|c|ccccc}
       \toprule 
       Exclude   & nAP & Bird & Bus & Cow & Motor & Sofa \\ 
       \midrule
       Bird      & 65.6 & {\R{33.0}}    & 84.5          & 70.7          & 77.9          & 61.6 \\
       Bus       & 64.1 & 56.9          & {\R{49.1}}    & 77.1          & 76.5          & 61.1 \\
       Cow       & 65.4 & 57.8          & 84.1          & {\R{46.1}}    & 77.0          & 62.2 \\
       Motor     & 67.5 & 60.6          & 83.9          & 73.9          & {\R{52.3}}    & 66.8 \\
       Sofa      & 65.7 & 62.4          & 83.5          & 75.8          & 74.9          & {\R{32.0}}\\
       \bottomrule
    \end{tabular}}
\end{table}

\begin{figure}[t]
    \vspace{-2mm}
    \centering
    \includegraphics[width=0.9\linewidth]{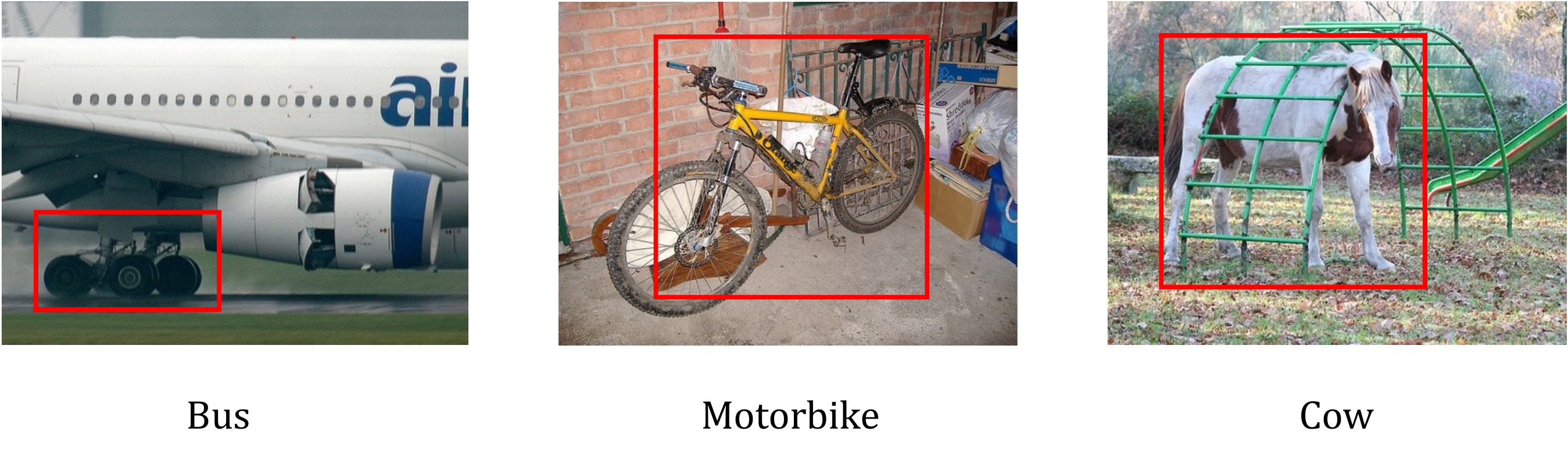}
    \vspace{-3mm}
    \caption{Examples of mined similar instances from excluded class dataset, \eg, ``wheel'' is mined for novel class ``bus'', ``bicycle'' is mined for novel class ``motorbike''}
    \label{fig:robost}
\end{figure}

\section*{Appendix B: Robustness of \emph{MINI}}

Although co-occurrence widely exists in benchmark datasets, there may be a case that the novel class does not co-occur with base classes.
In this section, we test the robustness of \emph{MINI} in such a case.
Specifically, we manually remove images that contain a selected novel class from the original base dataset of PASCAL VOC Novel Split1 with the Shot1 setting, and keep the novel dataset unchanged.
We then train a TFA and \emph{MINI} on this processed dataset, the results are shown in Table~\ref{tab:robustness}.
Surprisingly, even the base dataset does not contain the removed novel class, our \emph{MINI} can still significantly improve the performance for the excluded class, \eg, boost nAP50 by 5.9 (from 27.1 to 33.0) for ``Bird'' and ``9.7'' (39.4 to 49.1) for bus.
So what instances are mined by \emph{MINI} for these excluded novel classes? We draw some examples in Fig.~\ref{fig:robost}. We can see these mined novel instances share a strong texture or shape similarity with the exclude class, \eg, the wheel of the base class ``aeroplane'' is also a part for novel class ``bus'', the shape of the base class ``bicycle'' is similar to the novel class ``motorbike'', the texture of the base class ``horse'' is similar to the novel class ``cow''.
We conjecture learning from these similar instances of base classes can also promote the feature representation ability of the corresponding novel classes.

\begin{table*}[t]
    \caption{
        Ablation study for hyper-parameters of different components.
        Varying $\alpha$ and $N$ for adaptive thresholding in offline mining.
        Varying $\delta$ for online mining.
        Varying $\beta$ for IoU branching
    }
    \label{tab:hyper}
    \vspace{4mm}
    \centering
    \begin{tabular}{c|ccccc|c|ccccc}
        \toprule
                                  & \multicolumn{5}{c|}{nAP50}                                 &                           & \multicolumn{5}{c}{nAP50} \\ \midrule
        $\alpha$                  & 1          & 2         & 3         & 5         & 10        & $\delta$                  & 1         & 2         & 3         & 5         & 10          \\ \midrule
        \ 0.0              \      & \bd{63.6}  & \bd{68.4} & 66.7      & 69.6      & 67.7      & \ 0.5             \       & 19.0      & 23.7      & 13.4      & 14.2      & 18.4        \\
        \ 1.5              \      & 63.5       & 67.7      & \bd{66.8} & \bd{70.3} & \bd{68.8} & \ 0.7             \       & \bd{69.9} & \bd{72.5} & \bd{71.7} & \bd{72.7} & \bd{73.8}   \\ 
        \ 3.0              \      & 59.8       & 59.4      & 65.9      & 69.6      & 67.2      & \ 0.9             \       & 65.2      & 70.5      & 69.8      & 71.2      & 71.4        \\ 
        \midrule
        $N$                       & 1          & 2         & 3         & 5         & 10        &  $\beta$                 & 1         & 2         & 3         & 5         & 10          \\ \midrule
        \   150    \              & 62.4       & 65.9      & 66.0      & 68.9      & 67.7      & \ 0.5               \    & \bd{69.9} & 72.5      & \bd{71.7} & \bd{72.7} & \bd{73.8}   \\  
        \   300    \              & 63.5       & 67.7      & \bd{66.8} & \bd{70.3} & \bd{68.8} & \ 1.0               \    & 69.4	     & \bd{72.7} & 69.8	     & 34.1	     & 31.6        \\
        \   450    \              & \bd{63.9}  & \bd{68.3} & 66.3      & 68.8      & 67.3      & \ 2.0               \    & 69.1	     & 72.0	     & 30.4	     & 33.0	     & 72.3        \\ 
        \bottomrule
    \end{tabular}
    \vspace{-3mm}
\end{table*}

\section*{Appendix C: Hyper-parameters Ablation}

In \emph{MINI}, thera are 4 hyper-parameters introduced, $\alpha$ and $N$ for adaptive thresholding, $\delta$ for online mining, and $\beta$ for IoU branching.
Table~\ref{tab:hyper} analyzes the effect of different choices of hyper-parameters. 
When studying $\alpha$ and $N$, we do not involve the online mining mechanism and the fine-tuning; when studying $\delta$ and $\beta$, we do not involve the fine-tuning.
A smaller $\alpha$ and larger $N$ will lead to more kept mined novel instances, which is beneficial in lower shots, \eg, 1- and 2- shot, but can be harmful in higher shots since it may result in more false positives. We can observe the performance is not very sensitive to $\alpha$ and $N$, and we finally adopt $\alpha=1.5$ and $N=300$ for the offline mining. 
During online mining, it is necessary to set a relatively high $\delta$. Because a too-small $\delta$, \eg, $\delta=0.5$ can severely degrade the performance, as it will induce too many false positives to distract the learning of the student model.
And we found a large $\beta$ will disturb the training process, especially in higher shots. 
Through a coarse study, we adopt $\delta=0.7$ and $\beta=0.5$ for all experiments.

\begin{table}[t]
    \centering
    \caption{Performance comparison between Self-supervised Discriminative Model and Supervised Discriminative Model in offline mining. 
    For self-supervised model, we adopt MoCo v2~\cite{chen2020mocov2} w/ ResNet-50;
    For supervised model, we adopt a ResNet R-50~\cite{he2016deep} supervised trained on ImageNet as the counter-part
    }
    \label{tab:diff-ssl}
    \vspace{2mm}
    \begin{tabular}{c|ccccc}
        \toprule
        \multirow{2}{*}{Discriminative Model} & \multicolumn{5}{c}{nAP50}                                 \\ 
                                              & 1         & 2         & 3         & 5         & 10         \\ \midrule
         ImageNet Pre-train                   & 62.3      & 67.2      & 66.4      & \bd{70.3} & \bd{69.4}  \\
         MoCo v2                              & \bd{63.5} & \bd{67.7} & \bd{66.8} & \bd{70.3} & 68.8       \\
        \bottomrule
    \end{tabular}
\end{table}

\section*{Appendix D: Self-supervised Discriminative Model vs. Supervised Discriminative Model in Offline Mining}

The offline mining mechanism leverages a self-supervised discriminative model to collaboratively mine implicit novel instances with the trained FSOD network, but how about using a supervised-learned pre-trained model to replace the SSL model?
Table~\ref{tab:diff-ssl} shows the comparison between SSL model MoCo v2 and an ImageNet supervised pre-trained ResNet-50.
Overall the performance of the SSL model compares favorably against the supervised counterpart,
especially on lower shots, \eg, $K=1, 2, 3$; but slightly inferior in higher shots, \eg, $K=10$.
This demonstrates the solid discriminative ability of the SSL discriminative model for offline mining in \emph{MINI}.